\documentclass[11pt]{article}

\usepackage[final,main]{neurips_2025}

\usepackage{amsmath,amsfonts,amssymb}
\usepackage{bbm} 
\usepackage{graphicx}
\usepackage{booktabs}
\usepackage{url}

\title{teLLMe Why (Ain't Nothing but a Jam):\\
Exploratory Causal Analysis of Urban Driving Data}

\author{
Qiwei Li \quad Jorge Ortiz \\
Rutgers University, Department of Electrical and Computer Engineering \\
\texttt{qiwei.li, jorge.ortiz@rutgers.edu}
}

\begin{document}

\maketitle

\begin{abstract}
Traffic agencies now have access to large volumes of video-derived data for studying safety and congestion. Most of these data are observational and collected without interventions, which makes causal questions such as “How would rain change traffic density?” difficult to answer. We present teLLMe, a system for exploratory causal analysis of urban driving datasets. The system starts from a structured event table built from dashcam annotations and combines causal structure learning with the PC algorithm, bootstrap-based stability checks, and query-specific effect estimation using linear regression and DoWhy. Natural-language questions are mapped to structured causal queries through a schema-aware LLM, enabling users to specify treatments, outcomes, and subpopulations. teLLMe returns a “Causal Card” that summarizes effect estimates, adjustment sets, DAG support, and assumptions, followed by a short natural-language explanation. Case studies on BDD-derived traffic events show that the system can surface plausible relationships involving weather, peak hours, and traffic density, while making uncertainty and modeling choices explicit. The system is designed as a tool for hypothesis generation and expert reasoning rather than a source of definitive causal claims.
\end{abstract}

\section{Introduction}

Cities now generate large amounts of video through dashcams, CCTV, and mobile devices. 
These streams capture routine traffic behavior and environmental conditions, and they offer a 
rich source of information for studying safety and congestion. Analysts often want to ask causal 
questions about this data. For example, does rain reduce traffic density at urban intersections, 
or do peak-hour periods increase demand on specific corridors. Most video, however, is 
observational and collected without interventions, which makes it difficult to distinguish 
genuine causal effects from confounding and exposure differences.

Turning raw video into variables that support causal reasoning introduces several challenges. 
The underlying data are imbalanced and confounded: weather, time of day, scene type, and demand 
patterns interact in ways that are not controlled by design, and some combinations occur rarely. 
The analysis pipeline itself is multi-stage. Object detections must be aggregated into 
event-level features, a plausible causal structure among those variables must be inferred, 
and treatment effects must be estimated for specific questions. Any resulting estimates must also 
be communicated to practitioners who are not expected to interpret causal graphs or adjustment rules.

\begin{figure}
    \centering
    \includegraphics[width=1\linewidth]{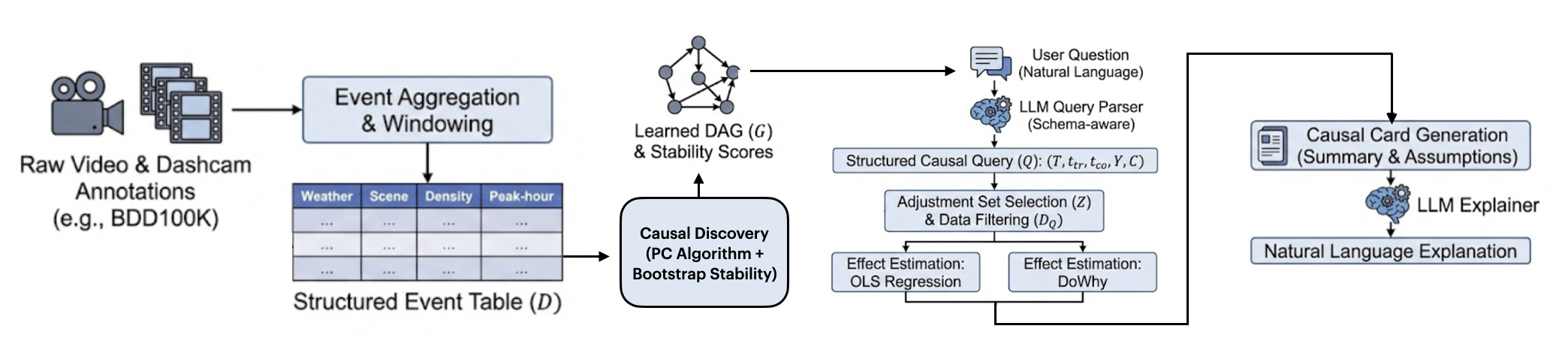}
    \caption{\small
        teLLMe has an offline phase that aggregates dashcam annotations into fixed-length
        windows, constructs a structured event table, and learns a causal graph using the
        PC algorithm with domain constraints and bootstrap stability. Online, a schema-aware
        LLM converts natural-language questions into structured causal queries specifying
        treatment, control, outcome, and filters. The system selects a backdoor adjustment
        set from the learned DAG, filters the event table, and estimates effects with linear
        regression and DoWhy. It returns a Causal Card with effect estimates, adjustment sets,
        DAG evidence, and assumptions, followed by a brief explanation.
        }
    \label{fig:placeholder}
\end{figure}

We introduce teLLMe, a system that connects these stages into a single workflow for exploratory 
causal analysis of urban driving data. The overall architecture of the system is shown in 
Figure~\ref{fig:placeholder}. teLLMe starts from a structured event table built from dashcam 
annotations and learns a causal graph over a curated set of variables using the PC algorithm with 
domain constraints and bootstrap resampling. Users pose questions in natural language. A schema-aware 
LLM translates each question into a formal causal query specifying treatments, outcomes, and 
subpopulation filters. teLLMe then selects a backdoor adjustment set based on the learned graph and 
estimates average treatment effects using linear regression and DoWhy. The results are presented in a 
Causal Card that reports effect estimates, adjustment sets, graph evidence, and key assumptions, 
followed by a brief natural-language explanation.

The contribution of teLLMe is the integration of these components into a query-driven system that 
makes causal assumptions explicit and keeps the workflow reproducible. The goal is to support the 
generation and inspection of plausible causal hypotheses from large video-derived datasets and to 
give domain experts a structured way to interrogate the evidence.

\section{Data and System Overview}
\label{sec:overview}

\subsection{Event dataset from dashcam annotations}

We work with a structured event table built from a large dashcam corpus such as BDD100K \citep{xu2017end}. 
Raw detections and metadata are aggregated into fixed-length windows, each represented by a row with 
weather labels, scene type, traffic density, peak-hour status, and basic temporal indicators 
(e.g., weekday/weekend and time-of-day bins). Traffic density is computed as the count of detected vehicles 
per minute. We drop windows with missing labels and clip extreme values to avoid outliers.
Because some combinations of weather and scene type are rare, we create a balanced subset using 
stratified sampling for the main analyses and retain the full dataset for sensitivity checks. 

\subsection{System architecture}

Figure~\ref{fig:placeholder} provides an overview of the system architecture. teLLMe is organized around two linked stages: an offline discovery phase that builds the structural substrate for causal reasoning, and an online query phase that turns user questions into structured causal analyses. The offline stage prepares the data and learns a causal graph with stability information, and the online stage parses user queries, selects adjustment sets, and estimates effects, producing a Causal Card for each query.

\paragraph{Offline.}
We standardize variables, learn a causal graph over selected features using the PC algorithm 
\citep{spirtes2000causation, kalisch2007estimating}, apply domain constraints to rule out implausible 
directions, and run bootstrap resampling to record edge stability. The resulting DAG and stability 
scores are stored for later use.

\paragraph{Online.}
Users provide a natural-language question. A schema-aware LLM converts it into a structured causal query 
with a treatment, outcome, and optional filters. The system selects a backdoor adjustment set based on 
the learned DAG, filters the event table accordingly, and estimates an average treatment effect using 
linear regression and DoWhy \citep{dowhy}. The output is a Causal Card that reports effect estimates, 
adjustment sets, graph evidence, and assumptions, along with a short natural-language explanation.

\section{Methods}
\label{sec:methods}

\subsection{Causal graph learning with PC and bootstrap}

Let $V$ denote a subset of event-level variables deemed relevant for causal analysis, such as
\[
V = \{\texttt{weather}, \texttt{is\_peak\_hour}, \texttt{traffic\_density}, \texttt{scene\_type}, \texttt{weekday}, \texttt{time\_bin}\}.
\]
We treat the event table as samples from a joint distribution over $V$ and aim to learn a directed acyclic graph (DAG) $G$ that encodes candidate causal relationships among these variables.

We use the PC algorithm~\citep{spirtes2000causation, kalisch2007estimating} as implemented in \texttt{pgmpy}. PC starts from a complete undirected graph over $V$ and iteratively removes edges based on conditional independence tests, then orients the remaining edges using a set of logical rules. We incorporate domain knowledge through \emph{forbidden edges} (e.g., disallowing edges from \texttt{traffic\_density} to \texttt{weather}) and, if desired, \emph{required edges}. All variables used for PC are discretized or encoded as needed for the chosen independence tests.

To quantify robustness, we perform a simple bootstrap stability analysis~\citep{spirtes2000causation, kalisch2007estimating}. We draw $B=20$ bootstrap resamples of the event table (with replacement), run PC on each resample, and record how often each directed edge appears. This yields a stability score $s(e) \in [0,1]$ for each edge $e$ in $G$, which we later expose in the Causal Cards. Intuitively, edges with high $s(e)$ are more stable under resampling, while edges with low $s(e)$ should be treated with caution.

\subsection{Query parsing and DAG-based adjustment selection}

Users interact with teLLMe through natural-language questions. To bridge between free text and the fixed dataset schema, we define a structured \emph{Causal Query}:
\[
Q = (T, t^{\mathrm{treated}}, t^{\mathrm{control}}, Y, C),
\]
where $T$ is the treatment variable, $t^{\mathrm{treated}}$ and $t^{\mathrm{control}}$ are two values of $T$, $Y$ is the outcome variable, and $C$ is a set of conditioning constraints (e.g., \texttt{scene\_type}~=~intersection, \texttt{is\_peak\_hour}~=~1).

We prompt an LLM with (i) the dataset schema, including variable names, types, and allowed categorical values, and (ii) the user question, and instruct it to output $Q$ in a constrained JSON format. The resulting specification is then validated against the schema: we enforce strict type checking, reject references to undefined variables, and discard invalid categorical values. If validation fails, the system returns a user-facing error and optionally falls back to a simple rule-based parser for a subset of templates. This schema-aware parsing layer is crucial for preventing hallucinated columns and keeping causal queries grounded in the actual dataset.

Given $Q$ and the learned DAG $G$, we select a backdoor adjustment set $Z$ for the treatment--outcome pair $(T,Y)$. In the simplest implementation, we start from a candidate list of covariates (e.g., \texttt{scene\_type}, \texttt{weekday}, \texttt{time\_bin}) and include those that are parents of $T$ or $Y$ in $G$ while avoiding descendants of $T$. This heuristic approximates a backdoor adjustment set~\citep{pearl2009causality} and encodes our modeling choice about which variables to condition on when estimating the effect of $T$ on $Y$.






\subsection{Effect estimation and Causal Cards}

For the filtered dataset $\mathcal{D}_Q$, we estimate an average treatment effect
\[
\tau_Q = \mathbb{E}[Y \mid do(T = t^{\mathrm{treated}}), C]
          - \mathbb{E}[Y \mid do(T = t^{\mathrm{control}}), C],
\]
using the assumptions encoded in $G$ and the selected adjustment set $Z$. 


\paragraph{Linear regression (OLS).}
We fit an ordinary least squares model
\[
Y = \alpha + \beta T + \gamma^\top Z + \varepsilon,
\]
with categorical variables in $Z$ one-hot encoded. The coefficient $\beta$ is taken as the
estimate of $\tau_Q$, and we report its point estimate, standard error, and confidence
interval.

\paragraph{DoWhy estimator.}
We also use DoWhy~\citep{dowhy} to construct a \emph{CausalModel} from
$\mathcal{D}_Q$, the treatment $T$, the outcome $Y$, and a DOT version of the
learned DAG $G$. The model identifies a backdoor estimand and estimates it via
linear regression, giving a second estimate of $\tau_Q$ under the same
adjustment set. DoWhy supports additional estimators, including
inverse-propensity and doubly robust methods, but we use linear regression for
consistency with the OLS baseline.

Optionally, we bootstrap $\mathcal{D}_Q$ to examine the spread of the estimated 
effect for the chosen estimator. This gives an empirical measure of stability 
beyond the usual standard errors.

\begin{figure}[t]
    \centering
    \includegraphics[width=0.55\linewidth]{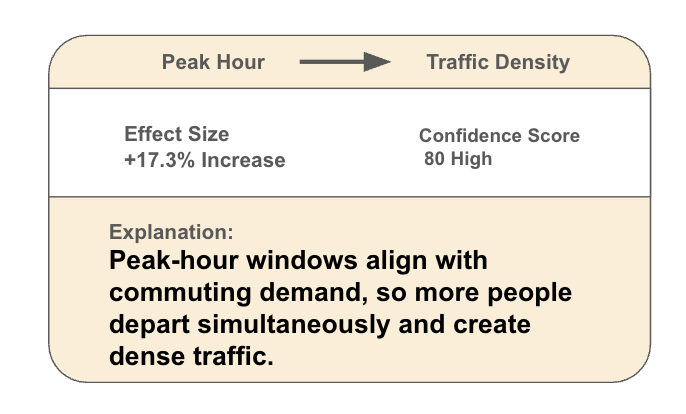}
    \caption{Example Causal Card summarizing the query, effect estimate, adjustment set, 
    DAG information, and a short explanation.}
    \label{fig:causal-card}
\end{figure}

\paragraph{Causal Cards.}
Each query is summarized in a compact Causal Card with the parsed variables, the 
effect estimate and confidence interval, treated/control sample sizes, any direct 
$T \to Y$ edge and its stability, and relevant caveats. A brief natural-language 
explanation is generated from these fields.

A short natural-language explanation is then generated from the card for 
non-technical users.

\begin{figure}[t]
    \centering
    \small

    \begin{tabular}{lccc}
        \toprule
        Estimator & ATE & 95\% CI & N (treated/control) \\
        \midrule
        OLS & 0.024 & [0.014, 0.034] & 4{,}268 / 6{,}154 \\
        \bottomrule
    \end{tabular}

    \vspace{1em}

    \centering
    
    \label{tab:outcome-dist}
    \begin{tabular}{crrrrrrrr}
        \toprule
        is\_peak\_hour & count & mean & std & p25 & median & p75 & min & max \\
        \midrule
        0 & 6154 & 0.3827 & 0.2466 & 0.20 & 0.35 & 0.50 & 0 & 1 \\
        1 & 4268 & 0.5291 & 0.2835 & 0.30 & 0.50 & 0.75 & 0 & 1 \\
        \bottomrule
    \end{tabular}

    \caption{ATE summary (top) and outcome distribution (bottom) for peak-hour 
    vs.\ off-peak on highway traffic density.}
    \label{fig:q2-ate}
\end{figure}




\section{Case Studies}
\label{sec:casestudies}

We illustrate teLLMe on several representative queries using the BDD-derived event dataset. 
All estimates and confidence intervals come from the specified subpopulations and adjustment sets.

\subsection{Weather and traffic density at urban intersections}

\begin{quote}
\emph{What is the effect of rainy versus clear weather on traffic density at urban intersections during peak hours?}
\end{quote}

For this query, the treatment is weather, with rainy as the treated 
condition and clear as the control. The outcome is traffic density, 
restricted to windows where scene type is urban intersection and 
the peak-hour indicator is one. The adjustment module selects time of day and 
total objects. In this run, the learned DAG does not contain a direct 
weather $\rightarrow$ traffic density edge.

Using OLS with the selected adjustment set, the estimated effect is $-0.036$ with a 
95\% confidence interval from $-0.047$ to $-0.024$, based on 1,840 rainy windows and 
7,681 clear windows.

\subsection{Peak-hour effects on highway traffic density}

\begin{quote}
\emph{How do peak-hour periods affect traffic density on highways under clear weather?}
\end{quote}

In this query, the treatment is \texttt{is\_peak\_hour} and the outcome is 
\texttt{traffic\_density}. The conditions restrict the data to 
\texttt{scene\_type}=\texttt{highway} and \texttt{weather}=\texttt{clear}. 
The adjustment module selects \texttt{time\_of\_day}, \texttt{weather}, and 
\texttt{total\_objects}. The learned DAG for this subpopulation does not contain a 
direct \texttt{is\_peak\_hour} $\rightarrow$ \texttt{traffic\_density} edge.

Using OLS with the selected adjustment set, the estimated effect is 0.024 with a 
95\% confidence interval from 0.014 to 0.034, based on 4,268 peak-hour windows 
and 6,154 off-peak windows.

\subsection{Sensitivity to adjustment and balancing choices}

We compare DAG-based adjustment with a fixed adjustment set and repeat analyses on the imbalanced 
dataset. Ignoring the DAG sometimes produces larger effects and narrower intervals, consistent with 
under-adjustment. Analyses on the imbalanced data occasionally produce more extreme estimates and 
greater uncertainty. These differences highlight the influence of adjustment and sampling choices; 
teLLMe surfaces these decisions directly in the Causal Cards.

\section{Discussion and Limitations}
\label{sec:discussion}

teLLMe works entirely with observational event data, so all effects depend on the 
assumptions encoded in the learned DAG and the selected adjustment sets. Several 
important factors—driver intent, road surface conditions, and weather severity—
are not observable in dashcam footage, which means the reported effects should be 
treated as plausible explanations rather than definitive causal claims.

The system treats event windows as independent and does not model temporal or 
spatial structure. Traffic patterns often depend on both, and capturing those 
dependencies would require different causal discovery methods and richer data.
Measurement limits also constrain the questions we can answer. Traffic density is 
straightforward to compute, but more safety-oriented surrogates, such as near-miss 
measures, are harder to derive reliably from video alone.
We have not yet evaluated the Causal Cards with practitioners. They are designed to 
make assumptions and uncertainty clear, but we do not yet know how analysts or 
planners interpret them or what forms of guidance they find most useful.

Beyond this standalone prototype, teLLMe is being developed as part of the Redddot project, a broader platform for participatory urban safety and mobility analytics. Redddot seeks to give planners, researchers, and community stakeholders access to interpretable views of heterogeneous urban data, including video-derived events, traffic indicators, and contextual information about places. Within this context, teLLMe plays the role of a causal reasoning and explanation module: it turns dashcam-derived event tables into queryable ``what-if'' analyses, and its Causal Cards can be surfaced alongside other Redddot views to show how candidate effects, uncertainty, and assumptions relate to specific locations and populations. This connection grounds teLLMe's design in a concrete application setting and highlights its potential to mediate human--AI collaboration around urban decisions.

The pipeline gives analysts and other Redddot stakeholders a direct way to test specific causal questions on video-derived data while keeping assumptions explicit. The resulting estimates are intended as inputs to further analysis and deliberation, not final answers.

\section*{Acknowledgements}
This work was supported by the National Science Foundation as part of the Center for Smart Streetscapes under Cooperative Agreement EEC-2133516 and by NSF Grant No. 2429672.

\bibliographystyle{plainnat}
\bibliography{references}

@book{pearl2009causality,
  title={Causality},
  author={Pearl, Judea},
  year={2009},
  publisher={Cambridge university press}
}

@book{spirtes2000causation,
  title={Causation, prediction, and search},
  author={Spirtes, Peter and Glymour, Clark N and Scheines, Richard},
  year={2000},
  publisher={MIT press}
}

@inproceedings{xu2017end,
    title={End-to-end learning of driving models from large-scale video datasets},
    author={Xu, Huazhe and Gao, Yang and Yu, Fisher and Darrell, Trevor},
    booktitle={The IEEE Conference on Computer Vision and Pattern Recognition (CVPR)},
    year={2017}
}

@article{kalisch2007estimating,
  title   = {Estimating high-dimensional directed acyclic graphs with the PC-algorithm},
  author  = {Kalisch, Markus and B{\"u}hlmann, Peter},
  journal = {Journal of Machine Learning Research},
  volume  = {8},
  number  = {Mar},
  pages   = {613--636},
  year    = {2007}
}

@article{dowhy,
  title   = {DoWhy: An End-to-End Library for Causal Inference},
  author  = {Sharma, Amit and Kiciman, Emre},
  journal = {arXiv preprint arXiv:2011.04216},
  year    = {2020}
}

\end{document}